\documentclass{article}
\usepackage{spconf,amsmath,graphicx,booktabs, amssymb}

\makeatletter
\let\NAT@parse\undefined
\makeatother
\usepackage[colorlinks, urlcolor=black, linkcolor=blue, citecolor=blue]{hyperref}

\usepackage{enumitem}
\setlist{nosep, leftmargin=14pt}

\usepackage{mwe} 

\title{Pyramid Attention Network for Medical Image Registration}
\name{Zhuoyuan Wang, Haiqiao Wang, Yi Wang$^{*}$
\thanks{$*$ Corresponding author}
\thanks{This work was supported in part by the National Natural Science Foundation of China under Grant 62071305, in part by the Guangdong-Hong Kong Joint Funding for Technology and Innovation under Grant 2023A0505010021, and in part by the Guangdong Basic and Applied Basic Research Foundation under Grant 2022A1515011241.}
}
\address{
	School of Biomedical Engineering,
	Shenzhen University, Shenzhen, China,\\
	and the Medical UltraSound Image Computing (MUSIC) Lab, Shenzhen, China,\\
	and the Smart Medical Imaging, Learning and Engineering (SMILE) Lab, Shenzhen, China\\}
\begin{document}
%
\maketitle
\begin{abstract}
The advent of deep-learning-based registration networks has addressed the time-consuming challenge in traditional iterative methods.
However, the potential of current registration networks for comprehensively capturing spatial relationships has not been fully explored, leading to inadequate performance in large-deformation image registration.
The pure convolutional neural networks (CNNs) neglect feature enhancement, while current Transformer-based networks are susceptible to information redundancy.
To alleviate these issues, we propose a pyramid attention network (PAN) for deformable medical image registration.
Specifically, the proposed PAN incorporates a dual-stream pyramid encoder with channel-wise attention to boost the feature representation.
Moreover, a multi-head local attention Transformer is introduced as decoder to analyze motion patterns and generate deformation fields.
Extensive experiments on two public brain magnetic resonance imaging (MRI) datasets and one abdominal MRI dataset demonstrate that our method achieves favorable registration performance, while outperforming several CNN-based and Transformer-based registration networks.
Our code is publicly available at~\textit{https://github.com/JuliusWang-7/PAN}.
\end{abstract}
\section{Introduction}
\label{sec:intro}
Deformable image registration has consistently stood as a pivotal concern in the field of medical image analysis.
Traditionally, algorithms like Elastix~\cite{klein2009elastix} and SyN~\cite{syn} have been proposed to iteratively solve the optimization problem to accomplish image registration.
Although these methods can yield satisfactory results, they are usually computationally expensive and time-consuming. 

In recent years, with the ongoing evolution of convolutional neural networks (CNNs) and the emergence of spatial Transformer networks~\cite{15spatial}, deep-learning-based approaches have emerged as the promising solution to the registration problem.
Balakrishnan~\textit{et al}.~\cite{19vxm} presented VoxelMorph, an Unet-based structure with a spatial transformer
network (STN), which achieved similar registration accuracy to traditional methods while offering fast inference speed.
Zhu~\textit{et al}.~\cite{zhu2021joint} proposed an end-to-end joint affine and deformable network for brain MRI registration.
Although above networks provided backbone architectures for the registration task, they did not pay too much attention on the feature enhancement, which might be useful for the enhancement of registration performance.

Since the successful application of the Transformer in the field of attention mechanisms, this structure has evolved into an immensely valuable method for enhancing features in the area of medical image registration.
Chen~\textit{et al}.~\cite{22transmorph} employed Transformer structures for the image registration, by adopting 3D Unet backbone and Swin-Transformer block as the part of encoder.
Moreover, Chen~\textit{et al}.~\cite{23transmatch} independently extracted features from each image with a Transformer-based backbone and utilized the Transformer's attention mechanism to match regions to these extracted features.
However, these one stage networks may not well handle complicated deformations.
To alleviate this issue, cascade and pyramid registration structures have been adopted to capture large and complicated deformations.
Zhao~\textit{et al}.~\cite{19rcn} proposed a cascaded architecture comprising multiple CNNs named recursive cascaded network (RCN).
But the repeated extraction of feature maps by RCN carries the potential risk of overfitting and challenges in propagating long-range information.
Kang~\textit{et al}.~\cite{kang2022dual} presented a dual-strem pyramid registration network (Dual-PRNet++) leveraging a two-stream 3D encoder-decoder network.
Cao~\textit{et al}.~\cite{cao2021edge} applied both edge information and original images into the pyramid network, guiding the network's attention towards edge details.
However, their oversight of different motion modes in low-resolution feature maps may mislead the subsequent generation of inaccurate deformation fields.
Zheng~\textit{et al}.~\cite{zheng2022residual} introduced a motion-aware architecture to aid the network in predicting diverse motion patterns.
Wang~\textit{et al}.~\cite{wang2023modet} leveraged the Transformer's inherent capacity to handle deformation estimation across multiple motion modalities.

In this study, we present an unsupervised pyramid attention network (PAN) for deformable medical image registration.
First, a pyramid encoder with channel-wise attention is employed to enhance the extracted features. 
Then, a multi-head local attention Transformer is designed as decoder to analyze motion patterns and generate deformation fields.
Moreover, an orthogonal regularization is employed to mitigate feature redundancy in the local attention Transformer, thereby learning more representative motion patterns.
Experiments on three public medical image datasets demonstrate the proposed PAN outperforms several CNN-based and Transformer-based registration networks.

\section{Method}
\label{sec:methods}

\begin{figure}[t]
\centering
\includegraphics[width=\columnwidth]{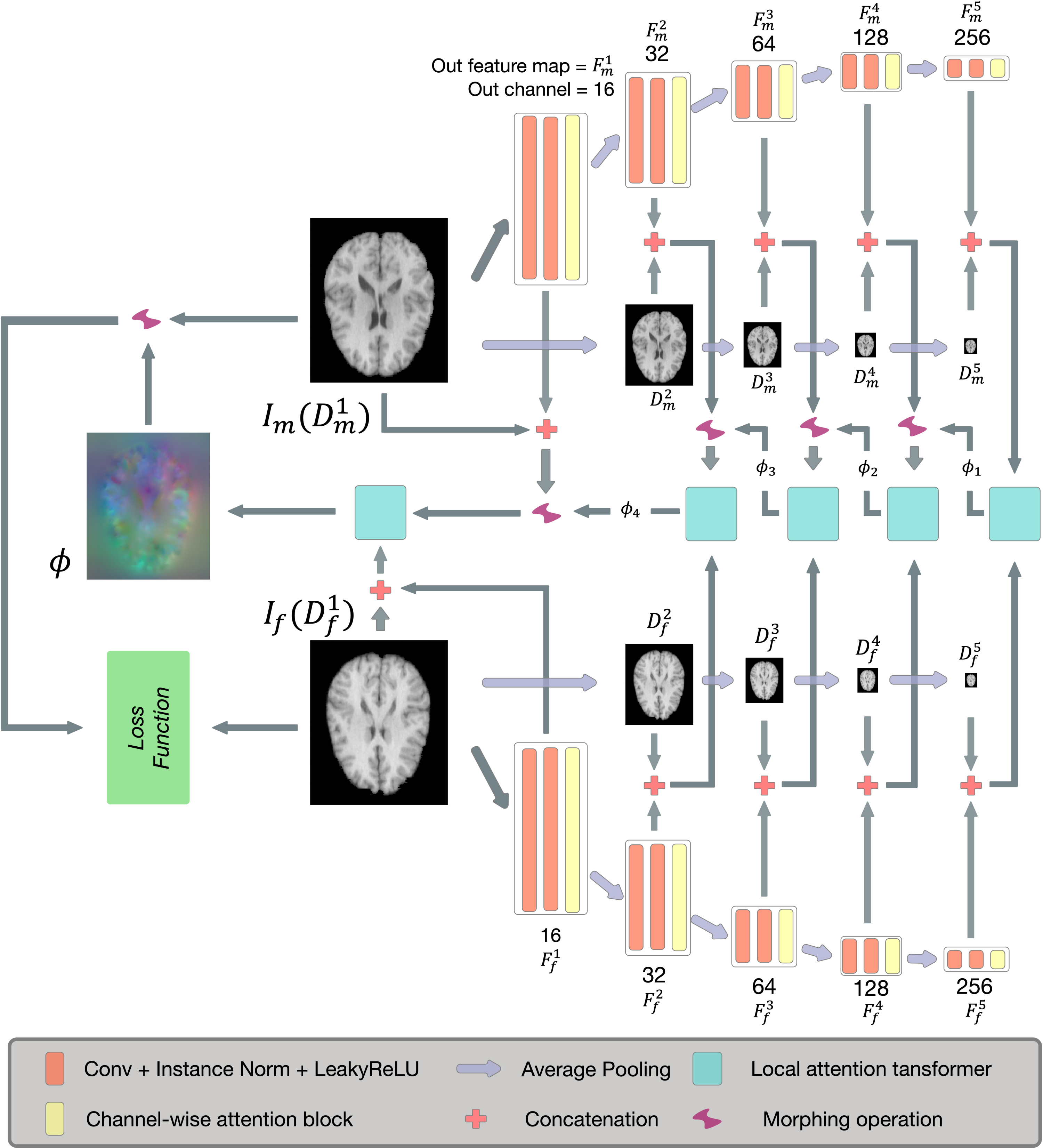}
\caption{The proposed pyramid attention network (PAN), which consists of the dual-stream pyramid encoder and the local attention Transformer decoder.}
\label{fig1}
\end{figure}

\subsection{Network Overview}
The proposed pyramid attention network is illustrated in Fig.~\ref{fig1}, which mainly consists of the dual-stream pyramid encoder and the local attention Transformer decoder.
The dual-stream weight-sharing pyramid encoder takes the moving image $I_m\in\mathbb{R}^{h \times w \times l}$ and the fixed image $I_f\in\mathbb{R}^{h \times w \times l}$ as input, and generates two sets of hierarchical features \{$F^1_m$, $F^2_m$, $F^3_m$, $F^4_m$, $F^5_m$\} and \{$F^1_f$, $F^2_f$, $F^3_f$, $F^4_f$, $F^5_f$\} using a series of channel-wise attention blocks.
The $h$, $w$, $l$ denote the image size in height, width and length, respectively.
Simultaneously, both $I_m$ and $I_f$ are downsampled to produce image sets \{$D^1_m$, $D^2_m$, $D^3_m$, $D^4_m$, $D^5_m$\} and \{$D^1_f$, $D^2_f$, $D^3_f$, $D^4_f$, $D^5_f$\}, then aligned with their corresponding feature maps.
In the decoder part, we leverage the local attention Transformer (LAT, see Fig.~\ref{fig2}) to interpret the feature maps and derive the deformation fields.
Initially, ($F^5_m$, $D^5_m$) and ($F^5_f$, $D^5_f$) are sent into the LAT module to analyze their motion patterns and yield the initial deformation field $\phi_1$.
$\phi_1$ is then utilized to deform $F^4_m$ and $D^4_m$, together with $F^4_f$ and $D^4_f$ as the input for the subsequent LAT module to generate $\phi_2$.
With the similar operations, $\phi_3$, $\phi_4$ and the final deformation field $\phi$ are generated.
Then $\phi$ is employed to deform $I_m$ (i.e., $I_m \circ \phi$) to get the registered image.
We follow VoxelMorph~\cite{19vxm}, using the normalized cross correlation $\mathcal{L}_{\operatorname{ncc}}$ to evaluate the image similarity, and the deformation regularization $\mathcal{L}_{\operatorname{reg}}$ to regularize the smoothness of the deformation field.
Furthermore, an orthogonal regularization loss $\mathcal{L}_{\operatorname{orth}}$ is used in the LAT module for the feature optimization, which will be described in Section~\ref{ortho_sec}.
Therefore, the total training loss will be:
\begin{equation}
\mathcal{L}_{train} = \mathcal{L}_{\operatorname{ncc}}(I_f, I_m \circ \phi) + \alpha\mathcal{L}_{\operatorname{reg}}(\phi) + \beta\mathcal{L}_{\operatorname{orth}},
\label{equloss} 
\end{equation}
where $\alpha$ and $\beta$ are the weighting factors.

\subsection{Channel-wise Attention Encoder}
Within each encoder unit, we employ two convolution blocks (each with a 3$\times$3$\times$3 convolution layer, an instance normalization layer~\cite{instancenorm}, and a LeakyReLU activation layer) and a squeeze-and-excitation (SE) block~\cite{seblock}.
Compared with the conventional convolution module that might overlook inter-channel relationships,
the SE block dynamically optimize the feature maps by incorporating global information.
We adopt such channel-wise attention mechanism to boost the encoder generating more representative features.

\subsection{Local Attention Transformer}
\label{lat_sec}
\subsubsection{The Architecture of LAT}
\begin{figure}[t]
	\centering
	\includegraphics[width=\columnwidth]{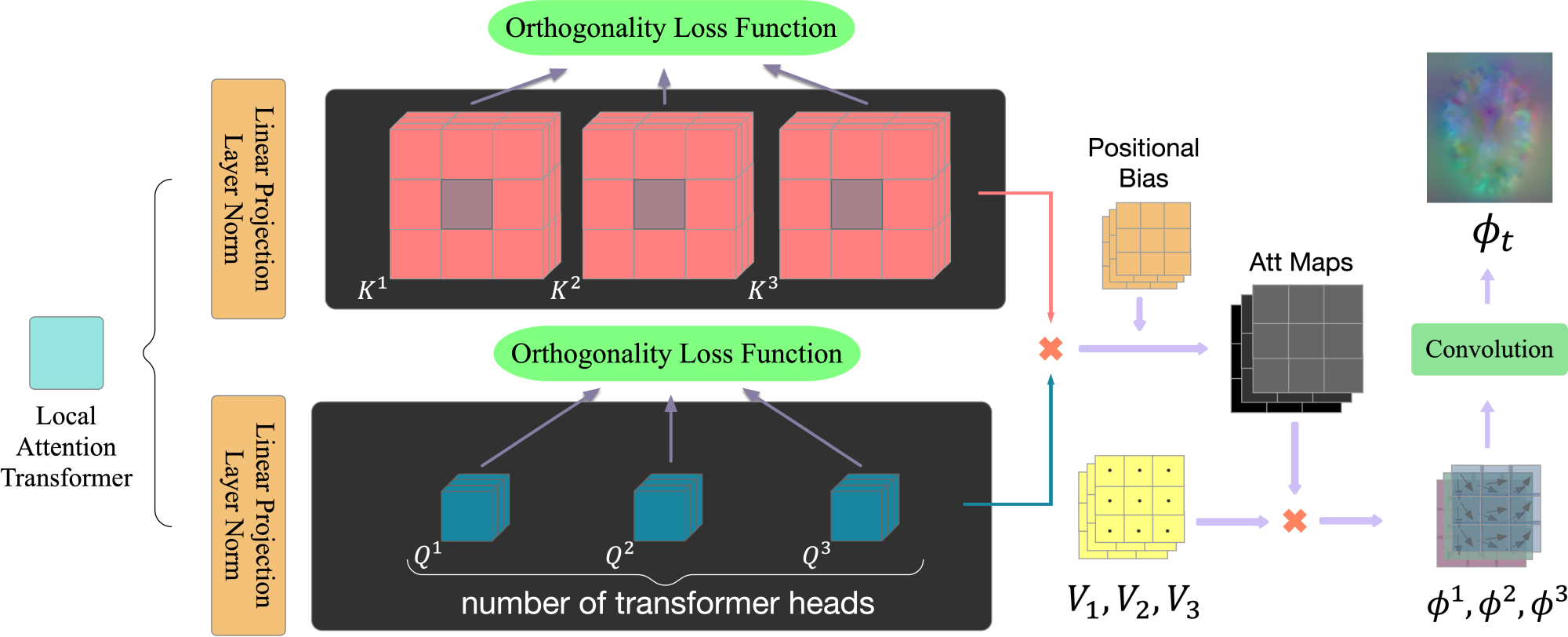}
	\caption{Illustration of the local attention Transformer (LAT). In this illustration, we take the attention head number $S$=3 for example.}
	\label{fig2}
\end{figure}

When handling volumetric data, traditional global self-attention tends to cause a huge rise in GPU memory with large-size inputs.
In contrast, the multi-head local attention mechanism mitigates this issue~\cite{23neighborhood}.
This mechanism localizes each pixel's attention scope to its nearest neighbors, resembling self-attention as its range expands while retaining translational equivalence.
We employ the local attention mechanism to analyze motion patterns and generate deformation fields, as shown in Fig.~\ref{fig2}.

We simply denote the moving feature maps, fixed feature maps, and their corresponding downsampled images from a specific level of the encoder as $F_m$, $F_f$, $D_m$ and $D_f$, respectively.
We separately concatenate ($F_f$, $D_f$) and ($F_m$, $D_m$) as the features in the channel dimension.
These concatenated features then pass a linear projection ($LP$) and a LayerNorm ($LN$)~\cite{ba2016layer} to yield the query ($Q$) and key ($K$), as represented in Equations~(\ref{equ1}) and~(\ref{equ2}):
\begin{equation}
Q = LN(LP(concat(F_f, D_f))), \label{equ1} \\
\end{equation}
\begin{equation}
K = LN(LP(concat(F_m, D_m))). \label{equ2} \\
\end{equation}
As shown in Fig.~\ref{fig2}, 
$Q \in\mathbb{R}^{S\times h \times w \times l \times \frac{c}{S}}$
and
$K \in\mathbb{R}^{S\times h \times w \times l \times \frac{c}{S}}$
can be separately divided by heads of the LAP to \{$Q^1$, $Q^2$,..., $Q^S$\} and \{$K^1$, $K^2$,..., $K^S$\},
where $S$ denotes the head number of the attention module,
$c$ is the channel number.
Then the local attention map of the $s$-th attention head can be calculated by:
\begin{equation}
LA^s = softmax(Q_x^s \cdot K_{N(x)}^s + P^s), \label{equ3} \\
\end{equation}
where $N(x)$ denotes the $n \times n \times n$ neighborhood for voxel $x$.
$P \in\mathbb{R}^{S\times n \times n \times n}$
is the learnable relative positional bias.
The local attention map $LA$ embeds the information of different motion patterns.
Therefore, we employ the $LA$ to weight the regular deformation field to generate a series of possible deformation subfields:
\begin{equation}
\phi^s = LA^s \cdot V \label{equ4},
\end{equation}
where the value ($V \in\mathbb{R}^{n \times n \times n}$) denotes the relative position coordinates for the neighborhood centroid.
Ultimately, we merge the deformation subfields \{$\phi^1$, $\phi^2$,..., $\phi^S$\} from each attention head through a convolutional layer to generate the deformation file $\phi_t \in\mathbb{R}^{h \times w \times l \times 3}$ at $t$-th decoding level.

\subsubsection{Orthogonal Regularization}
\label{ortho_sec}
To address the potential homogeneity in the content learned by the multi-head LAT,
we employ the orthogonality loss function to regularize the feature learning~\cite{brock2016neural}.
This loss ensures that the learning content diverges among different attention heads, thereby enhancing the analyzing of different motion patterns.
To achieve this, we first transform $Q$ and $K$ into flat vectors in terms of the dimension of LAT's heads, obtaining the $\mathcal{W} \in\mathbb{R}^{S\times u}$ ($u=h\cdot w\cdot l\cdot \frac{c}{S}$).
After acquiring this matrix $\mathcal{W}$, we normalize the vector represented by each head, which is then multiplied by its transposed matrix.
Then the orthogonality loss $\mathcal{L}_{\operatorname{orth}}$ can be calculated by:
\begin{equation}
\mathcal{L}_{\operatorname{orth}} = \mathcal{L}_{\operatorname{MSE}}(Norm(\mathcal{W})\cdot Norm(\mathcal{W})^T, \mathbb{I}),
\end{equation}
where $\mathbb{I}$ is an identity matrix, and $\mathcal{L}_{\operatorname{MSE}}$ is the computation of the mean square error (i.e., l2-norm).

\section{Experiments}
\label{sec:experiments}
\subsection{Datasets}
Experiments were conducted on two public brain MRI datasets (IXI~\cite{22transmorph} and LPBA40~\cite{lpba}), and one public abdominal MRI dataset (Learn2Reg AbdomenMRCT~\cite{learn2reg})\footnote{Please refer to the relevant references~\cite{22transmorph, lpba, learn2reg} for the detailed acquisition information for these public datasets.}.\\
\noindent \textbf{The IXI dataset} contains 576 brain MRI scans, each with 45 manually annotated region-of-interests (ROIs).
All MRI scans were cropped to the size of $160\times192\times224$.
461 scans (461$\times$460 pairs) were used for training and the remaining 115 scans (115$\times$114 pairs) were used for testing.\\
\noindent \textbf{The LPBA40 dataset} contains 40 brain MRI scans, each with 56 labeled ROIs.
All MRI scans were resampled to the size of $160\times192\times160$.
30 scans (30$\times$29 pairs) were employed for training and 10 scans (10$\times$9 pairs) were used for testing.\\
\noindent \textbf{The AbdomenMR dataset} contains abdominal MRI scans with 4 annotated organs.
Each MRI scan has the size of $192\times160\times192$.
38 scans (38$\times$37 pairs) were used for training and 10 scans (10$\times$9 pairs) were used for testing. 

\subsection{Implementation Details}
We deployed our method using PyTorch on a NVIDIA Tesla A100 GPU equipped with 40GB memory.
The neighborhood size $n$ was set to 3.
The number of attention heads in LAT were set to 8, 4, 2, 1, 1, from deep to shallow, and each head was equipped with 6 channels.
The weighting factors ($\alpha$, $\beta$) were set to (1, 1) and (0.5, 1) for brain and abdomen scans, respectively.
The training was conducted using Adam optimizer with a learning rate of 0.0001.
The batch size was 1.
Our code is publicly available at~\textit{https://github.com/JuliusWang-7/PAN}.

\subsection{Comparison Methods and Evaluation Metrics}
To demonstrate the efficacy of the proposed PAN, we compared several cutting-edge registration methods:
(1) SyN~\cite{syn}: a classical iterative registration method.
(2) VoxelMorph (VM)~\cite{19vxm}: a popular CNN-based single-stage registration model.
(3) Dual-PRNet++ (DPR++)~\cite{kang2022dual}: a CNN-based pyramid registration model.
(4) TransMorph (TM)~\cite{22transmorph}: a Transformer-based registration model.
(5) Vit-V-net (Vit-V)~\cite{vitvnet}: a CNN-Transformer-based registration model.

To quantify the performance of the registration methods, we computed the 3D metrics of Dice similarity coefficient (DSC), average symmetric surface distance (ASSD), and the Jacobian determinant of the deformation field ($|J(\phi)|<0$)~\cite{wang2023modet}.
The DSC is used to evaluate the region similarity between the fixed and registered images.
The ASSD is employed to evaluate the boundary similarity.
The Jacobian determinant is to assess the smoothness of the deformation field.
Higher value for DSC, lower values for ASSD and Jacobian determinant indicate better registration performance.

\begin{table}[t]
	\centering
	\caption{\label{jlab1}Quantitative results of different methods on IXI dataset (45 ROIs).}
	\footnotesize
	\begin{tabular}{@{}l|c|c|c}
		\toprule
		Methods & DSC (\%) & ASSD & $\%|J(\phi)|<0$\\
		\midrule
		Initial & 50.8$\pm$5.1$^*$ & 1.53$\pm$0.32$^*$ & / \\
		\midrule
		SyN~\cite{syn} & 64.9$\pm$3.5 & 0.89$\pm$0.20 & $<$0.0006\% \\
		VM~\cite{19vxm} & 62.0$\pm$4.1$^*$ & 1.01$\pm$0.23$^*$ & $<$2\% \\
		DPR++~\cite{kang2022dual} & 63.6$\pm$3.8$^*$ & 0.94$\pm$0.21$^*$ & $<$1\% \\
		Vit-V~\cite{vitvnet} & 62.8$\pm$4.0$^*$ & 0.98$\pm$0.22$^*$ & $<$2\% \\
		TM~\cite{22transmorph} & 63.9$\pm$3.9$^*$ & 0.93$\pm$0.21$^*$ & $<$2\% \\
		\midrule
		PAN (Ours) & \textbf{65.1$\pm$3.4} & \textbf{0.89$\pm$0.20} & $<$\textbf{0.0003\%} \\
		\bottomrule
	\end{tabular}\\
\end{table}

\begin{table}[t]
	\centering
	\caption{\label{jlab2}Quantitative results of different methods on LPBA40 dataset (56 ROIs).}
	\footnotesize
	\begin{tabular}{@{}l|c|c|c}
		\toprule
		Methods & DSC (\%) & ASSD & $\%|J(\phi)|<0$\\
		\midrule
		Initial & 54.1$\pm$4.6$^*$ & 2.60$\pm$0.30$^*$ & / \\
		\midrule
		SyN~\cite{syn} & 71.0$\pm$1.6 & 1.52$\pm$0.10 & $<$\textbf{0.00001\%} \\
		VM~\cite{19vxm} & 62.8$\pm$3.1$^*$ & 1.96$\pm$0.19$^*$ & $<$1\% \\
		DPR++~\cite{kang2022dual} & 68.7$\pm$1.8$^*$ & 1.65$\pm$0.12$^*$ & $<$0.1\% \\
		Vit-V~\cite{vitvnet} & 63.2$\pm$3.3$^*$ & 1.90$\pm$0.20$^*$ & $<$2\% \\
		TM~\cite{22transmorph} & 65.6$\pm$2.8$^*$ & 1.82$\pm$0.17$^*$ & $<$1\% \\
		\midrule
		PAN (Ours) & \textbf{71.1$\pm$1.6} & \textbf{1.51$\pm$0.10} & $<$0.0001\% \\
		\bottomrule
	\end{tabular}\\
\end{table}

\begin{table}[t]
	\centering
	\caption{\label{jlab3}Quantitative results of different methods on AbdomenMR dataset (4 ROIs).}
	\footnotesize
	\begin{tabular}{@{}l|c|c|c}
		\toprule
		Methods & DSC (\%) & ASSD & $\%|J(\phi)|<0$ \\
		\midrule
		Initial & 57.1$\pm$8.9$^*$ & 14.59$\pm$3.86$^*$ & / \\
		\midrule
		SyN~\cite{syn} & 67.4$\pm$15.2 & 13.93$\pm$9.36 & \textbf{$<$0.01\%} \\
		VM~\cite{19vxm} & 68.8$\pm$7.1$^*$ & 11.88$\pm$4.37$^*$ & $<$20\% \\
		DPR++~\cite{kang2022dual} & 70.1$\pm$8.3$^*$ & 10.96$\pm$4.31 & $<$3\% \\
		Vit-V~\cite{vitvnet} & 69.4$\pm$7.2$^*$ & 11.49$\pm$4.34$^*$ & $<$20\% \\
		TM~\cite{22transmorph} & 71.8$\pm$6.0 & 11.10$\pm$4.17$^*$ & $<$20\% \\
		\midrule
		PAN (Ours) & \textbf{73.1$\pm$8.6} & \textbf{10.26$\pm$4.49} & $<$0.03\% \\
		\bottomrule
	\end{tabular}\\
\end{table}

\subsection{Experimental Results}

\begin{figure}[t]
	\centering
	\includegraphics[width=0.95\columnwidth]{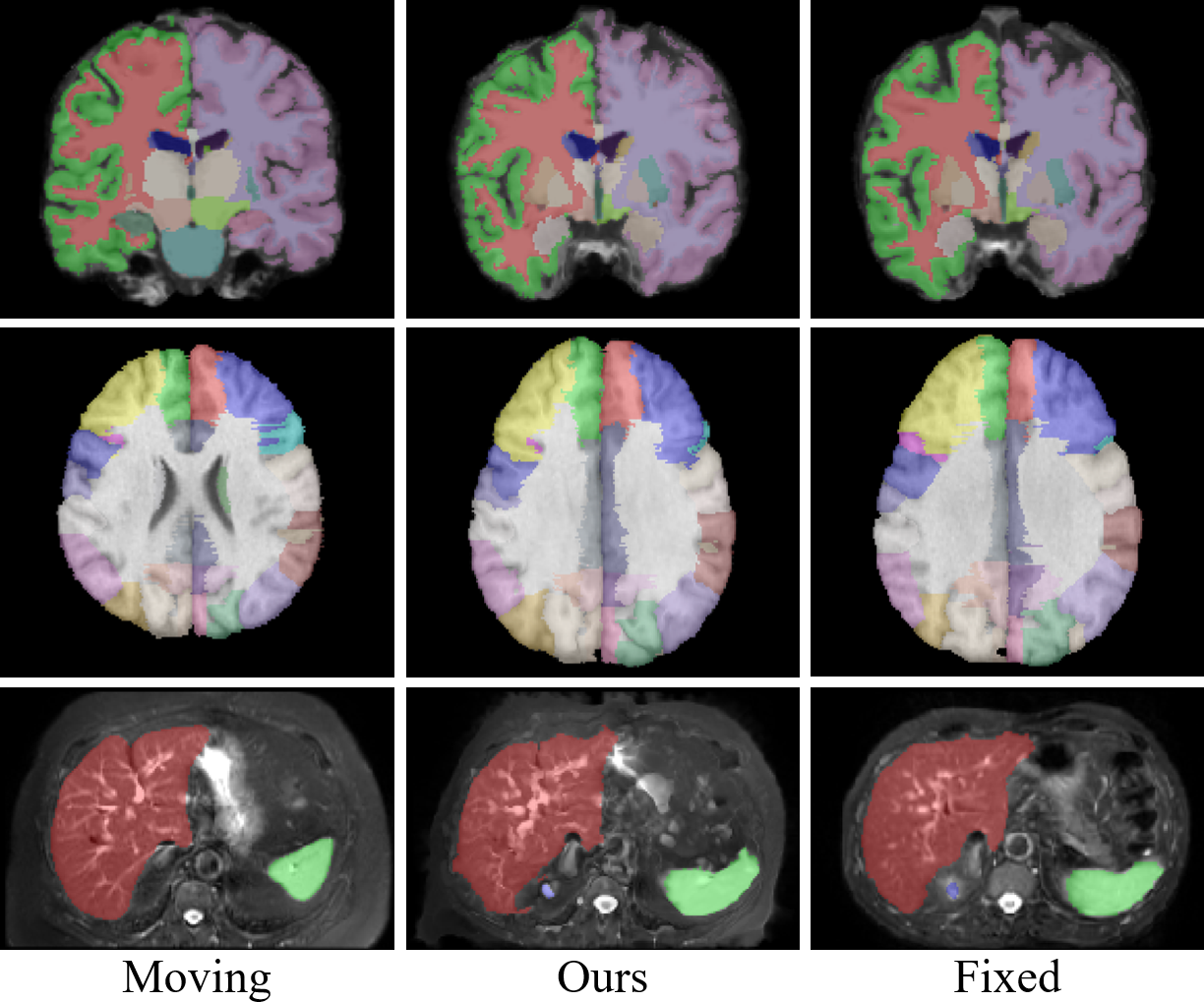}
	\caption{The visualized registration results. The colored regions indicate the masks of different anatomical structures. $1^{st}$ row: IXI; $2^{nd}$ row: LPBA40; $3^{rd}$ row: AbdomenMR.}
	\label{fig3}
\end{figure}

The quantitative results of different methods on the three testing datasets are shown in Tables~\ref{jlab1},~\ref{jlab2} and~\ref{jlab3}.
In these Tables, the symbol $^*$ indicates the DSC/ASSD results are statistically different from ours (Wilcoxon tests, \textit{p}$<$0.05).
It can be observed that our proposed PAN outperforms the majority of other deep-learning-based registration networks with respect to the accuracy metrics of DSC and ASSD.
By further observing the percentage of voxels with non-positive Jacobian determinant ($\%|J(\phi)|<0$),
it is obviously our method generates most smoothness deformation fields among all deep-learning-based models.
It is worth noting that all deep-learning-based models have much faster registration speed compared to the traditional method SyN.
Our method averagely takes 0.6 second to register a pair of volumes.

Fig.~\ref{fig3} shows some examples of the registered images from the three datasets.
Although the deformations between the fixed and moving images are large and complicated,
our method successfully matches corresponding anatomical structures and generates accurate registration results.

\section{Conclusion}
\label{sec:conclusion}
We present an unsupervised pyramid attention network (PAN) for deformable registration.
A multi-head local attention Transformer (LAT) is designed to analyze motion patterns and generate deformation fields.
An orthogonal regularization strategy is introduced to mitigate feature redundancy and learn more representative motion patterns in the LAT.
The designed LAT modules are integrated into the coarse-to-fine registration framework in the decoder, to capture long-range dependencies between feature maps.
Experimental results on three public medical image datasets have proven the superior performance of our proposed method.
Nonetheless, the capability of the PAN for registering images with large deformation still needs to be boosted.
One potential aspect to tackle this problem could be employing the recursive deformation estimation strategies~\cite{19rcn, 10423043}.\\

\noindent\textbf{Compliance with Ethical Standards}\\
This research study was conducted retrospectively using medical imaging data made available in open access by~\cite{22transmorph},~\cite{lpba} and~\cite{learn2reg}.
Ethical approval was not required as confirmed by the license attached with the open access data.

\hspace*{\fill}\\

\noindent\textbf{Conflicts of Interest}\\
The authors have no conflicts to disclose.

\bibliographystyle{IEEEbib}
\bibliography{refs}

\end{document}